\documentclass[lettersize,journal]{IEEEtran}

\usepackage{amsmath}
\usepackage{amsfonts}
\usepackage{array}

\usepackage{graphicx}
\usepackage{stfloats}
\usepackage{booktabs}
\usepackage{tabularx}
\usepackage{longtable}
\usepackage{ragged2e}
\usepackage{tikz}

\usepackage[caption=false,font=normalsize,labelfont=sf,textfont=sf]{subfig}

\usepackage{algorithm}
\usepackage{algpseudocode}

\usepackage{newfloat}
\usepackage{listings}
\usepackage[table,xcdraw]{xcolor}
\usepackage{color}

\usepackage{textcomp}
\usepackage{url}
\usepackage{xspace}
\usepackage{pifont}
\usepackage{mdframed}
\usepackage{cite}

\usepackage{listings}
\usepackage{fvextra}

\usepackage{makecell}

\lstdefinelanguage{PDDL}{
  basicstyle=\ttfamily\small,
  keywords={:action, :parameters, :precondition, :effect, and, when, forall, assign, increase},
  keywordstyle=\color{blue}\bfseries,
  identifierstyle=\color{black},
  sensitive=true,
  comment=[l]{;},
  commentstyle=\color{gray}\itshape,
  morestring=[b]",
  stringstyle=\color{orange}
}

\definecolor{qline}{HTML}{1F6FEB}
\definecolor{qbg}{HTML}{EEF5FF}
\definecolor{aline}{HTML}{2DA44E}
\definecolor{abg}{HTML}{EEF9F0}
\definecolor{brown}{rgb}{0.6, 0.4, 0.2}

\mdfdefinestyle{QAcommon}{
  linewidth=0.8pt,
  roundcorner=6pt,
  innerleftmargin=8pt,
  innerrightmargin=8pt,
  innertopmargin=6pt,
  innerbottommargin=6pt,
  skipabove=0pt,
  skipbelow=0pt
}

\mdfdefinestyle{Qstyle}{
  style=QAcommon,
  linecolor=qline,
  backgroundcolor=qbg
}

\mdfdefinestyle{Astyle}{
  style=QAcommon,
  linecolor=aline,
  backgroundcolor=abg
}

\newcommand{\sca}{\textit{SafeChat}\xspace}

\newcommand{\lh}{\textit{Lighthouse}\xspace}

\begin{document}

\title{\lh: On Building a
Trustworthy Digital Assistant for Exploring History with Civil Rights Archives}

\author{John A. Aydin, Angelica Williams, Steven Dindl, Nathan Saunders\\ Kate Boyd, Vandana Srivastava, Biplav Srivastava%
\thanks{All authors are with the University of South Carolina, Columbia, SC, USA.}%
\thanks{Manuscript received Month DD, YYYY; revised Month DD, YYYY.}}

\markboth{IEEE Transactions on X,~Vol.~X, No.~X, Month~Year}%
{Contributors: Lighthouse}

\maketitle

\begin{abstract}
History — the systematic study of change over time — employs methods from a variety of disciplines to study societal and community development. Historical studies provide insight into group dynamics and human behavior while also serving as a primary tool by which groups create mutual understanding and weigh options when deciding among courses of action.
Artificial intelligence (AI), as a decision-support aid, provides new tools for studying history effectively using multimodal data, including text, images, and audio. We present, to our knowledge, the first digital assistant named {\bf \lh}---commonly called a {\em chatbot}---for exploring the Civil Rights period of the United States using data from South Carolina. From an AI perspective, the system’s novelty lies in its guiding design principles of (a) focusing on trustworthiness of content and (b) by modeling and reducing potential harm to users from generated answers. It implements these with a neuro-symbolic approach that combines neural learning through a large language model (LLM) with rule-based symbolic solving, balancing risk of answers with generality of coverage. We evaluate \lh with both factual ({\em objective}) questions that admit a single acceptable answer as well as  {\em subjective} questions. 
For the latter, a dataset of subjective question-answer pairs (QAs) is created by discovering a ``gold standard'' by aggregating answers from independent experts. \lh has access to over 18k primary sources from a range of collections and can be easily extended.


\end{abstract}

\section{Introduction}



\begin{quote}
    “Those who cannot remember the past are condemned to repeat it.” – George Santayana, The Life of Reason, 1905.
\end{quote}

Libraries and archives have been digitizing and making their collections of primary resources available online through traditional content management and information retrieval systems for more than 30 years\cite{saa_american_memory}. Historians, students, and general audiences have adapted their learning to study these materials online, using accurate keywords and Boolean search mechanisms to retrieve relevant documents for their research. However, twentieth-century archives, such as Civil Rights collections, are large and include an abundance of materials, making the successful search more challenging. 

Digital assistants powered by AI, commonly referred to as {\em chatbots} or conversational assistants, have the potential to dynamically change how people interact effectively with digital collections in a {\em personalized } manner.
These digital assistants allow for more nuanced knowledge inquiry using natural language instead of precise terms and logical (e.g., Boolean) operators seen in traditional {\em information retrieval} (IR) systems like search engines\cite{conversation-ir-framework}.  This has the potential to open digital collections to {\em new questions and searches that the user hitherto did not know how to ask}. Instead of inputting search words such as “voting rights” and having 200 documents returned to then read through, a researcher could pose a more {\em natural} question, “How did people in South Carolina fight for voting rights?”  or using {\em different, but semantically similar, terms} like “How did people in South Carolina fight to have a voice in elections?” 
These in turn could elicit a direct response from a chatbot that discusses the findings from the archives with links to the relevant primary resources, providing context for the results. Such a natural interaction would allow for easier reviewing of results and narrowing or broadening of the search, instead of trying new search terms and essentially starting over each time.

However, having a personalized, but inherently sequential, form of interaction does pose new concerns about trust in chatbot responses and ensuring that the conversation and results are accurate, unbiased, and acceptable. 
This can be especially problematic with chatbots based on Large Language Models (LLMs), which build large statistical models and generate outputs based on probabilistic distributions, as their responses can include confident but inaccurate information, often referred to as hallucinations \cite{bang2025hallulens}.
Moreover, archives, as historical collections, may contain conflicting perspectives that require careful consideration and contextualization when being studied. Finally, users, including historians and researchers, need to have assurances that the chatbot will not give false or misleading information.

\begin{figure*}
    \centering
    \includegraphics[width=0.9\linewidth]{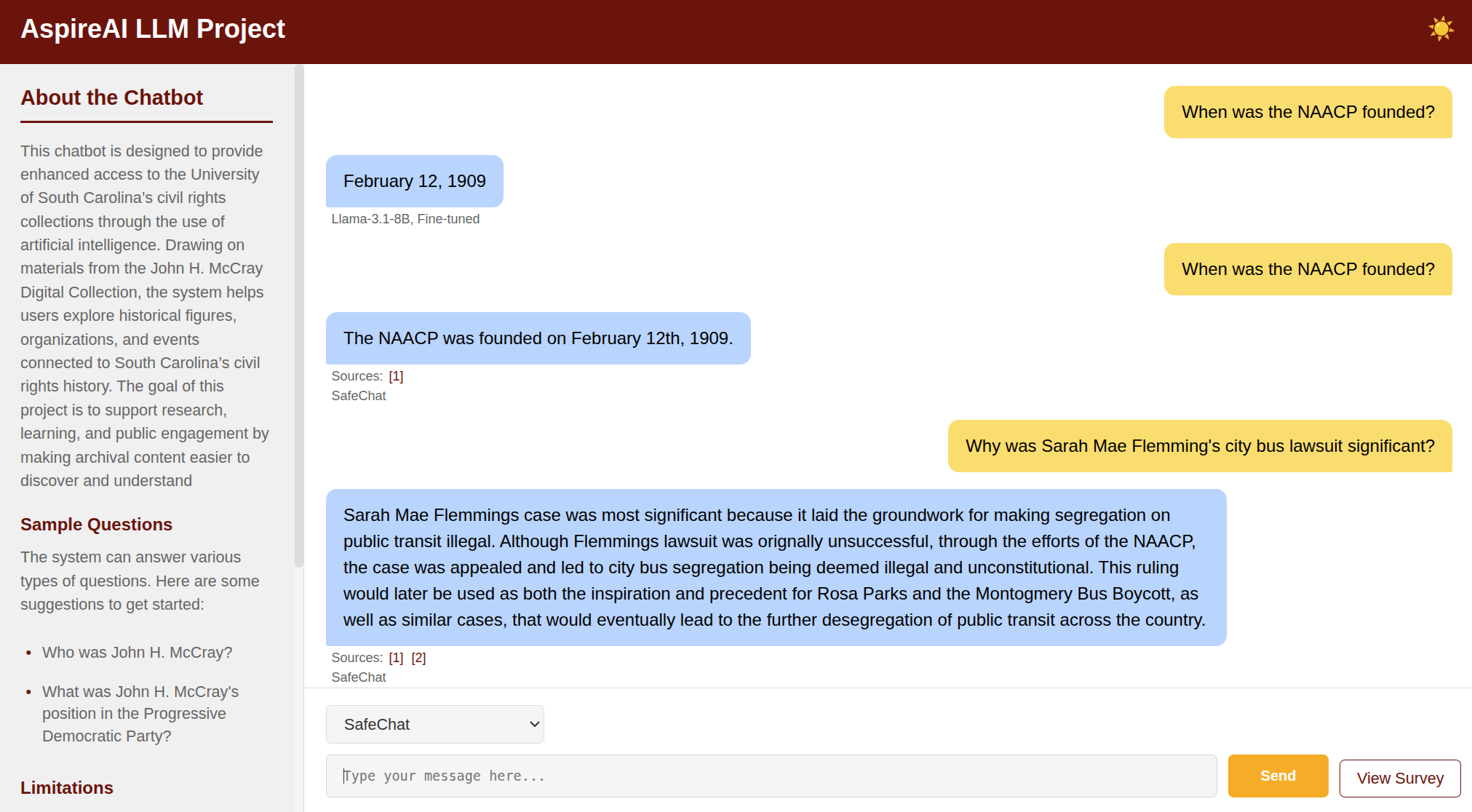}
    \caption{Example user interaction with \lh. The figure shows both S1 (LLM-based: Llama-3.1-8B, Fine-tuned) and S2 (rule-based: \sca) dialog managers - chatbot subcomponents - responding to user queries. S1 can provide an exact answer but does not include cited sources. S2 provides an expert-vetted answer with citations. \lh will {\bf dynamically learn} to call S1 and S2 based on potential benefits and risk of conversation.} 
    \label{fig:sgc_demo}
\end{figure*}

This project aims to explore how \textit{digital assistants} can improve upon the traditional search functionality of digital archives. Historians, librarians, and computer scientists are working together with the Civil Rights digital collections of the University of South Carolina to develop a trustworthy system, re-imagining how any researcher can interact with modern archives and library collections online. If successful, this project could change how researchers, students, and the public engage with complex historical collections. Figure~\ref{fig:sgc_demo} shows a demonstration of \lh allowing the user to ask objective and subjective questions and also change the backend to investigate how responses  are generated using learning or rule-based methods (dialog policy). 

The contributions of this paper are as follows: (1)  present and discuss issues pertaining to the deployment of artificial intelligence in the domain of historical archives, (2)  present \lh, a trustworthy digital assistant for natural-language exploration of Civil Rights archival collections, (3) compare LLM-based, rule-based, and hybrid neuro-symbolic approaches for trustworthy archival question answering. The remainder of this paper is organized as follows. We start by reviewing 
background and related work on Civil Rights archives and trustworthy conversational assistants, followed by formalizing the problem. 
We then describe our data processing, annotation, and AI approaches, followed by our evaluation. We close with a discussion of significance, limitations, and future work before concluding.

\label{sec:case-study} 


\section{Background \& Related Work}


\label{sec:related-work}

In this section, we provide background on conversational assistants, South Carolina Civil Rights history, and trustworthy AI. This will help contextualize the problem and approach.

\subsection{Conversational Assistants }

Chatbots, or conversational assistants, are software systems designed to interact with users through natural-language dialog. Unlike traditional information retrieval interfaces, which typically require users to formulate search queries and inspect retrieved results, conversational assistants can support contextual, multi-turn interactions in which users ask questions, refine requests, and receive synthesized responses. 


Early chatbots primarily relied on rule-based approaches. Given a user query, these systems matched keywords, patterns, or intents to a fixed set of predefined responses or actions. This paradigm was initially popularized by ELIZA, which used keywords in user inputs to select scripted responses that sustained the conversation \cite{ELIZA}. Later systems incorporated larger rule bases, variables, dialog state, and more sophisticated intent-recognition \cite{anatomy_of_ALICE_2009,bocklisch2017rasaopensourcelanguage}. We refer to these systems as \emph{answer-closed}: their outputs are constrained to responses, rules, or actions that have been explicitly authored, vetted, or configured prior to deployment. As a result, they can provide predictable behavior within their designed scope, but they are limited in their ability to answer novel questions or adapt to unanticipated user needs.


More recently, large language models (LLMs) have become a dominant foundation for chatbot development. These systems are often described as generative AI (genAI) because they construct responses dynamically at runtime rather than selecting from a fixed response set or executing manually specified dialog rules. This generative capability allows LLM-based chatbots to handle a wider range of open-ended requests, including explanation, summarization, and code generation \cite{srivastava2023beyond}. Their flexibility has contributed to rapid adoption in domains where users benefit from interactive assistance and where occasional errors can be tolerated or mitigated, such as language learning or software development \cite{henry_duolingo_ai_lessons_2023, stackoverflow2024devsurveyai}. LLM-based chatbots can suffer from hallucinations and require substantial training data, compute, and energy resources, unlike symbolic systems.


LLM-based chatbots are typically developed in two broad phases. The first is pre-training, a computationally intensive process in which a base model is trained on large-scale text corpora to learn general linguistic, factual, and reasoning patterns \cite{zhao2026surveylargelanguagemodels}. The second is post-training, which follows pre-training and aims to shape the model's behavior for practical use. In contrast to pre-training, which is primarily compute intensive, post-training is resource intensive; it requires high quality data, expert annotators, or human feedback. These resources are used to align the model with human preferences, improve task-specific performance, increase helpfulness and safety, or otherwise modify how the model responds in conversational settings \cite{zhang2025instructiontuninglargelanguage, ouyang2022traininglanguagemodelsfollow}.

The creation of domain-specific chatbots is commonly approached through either model adaptation or retrieval-based augmentation. In the first approach, a pre-trained model is further trained on domain-specific data, such as medical dialogs, customer support interactions, or legal documents. This process, commonly referred to as fine-tuning, adapts the model by updating some or all of its parameters, thereby internalizing domain-specific behavior or knowledge in the model weights \cite{hu2021loralowrankadaptationlarge}.
 In contrast, retrieval-augmented generation (RAG) leaves the model weights unchanged at inference time and instead conditions the model on external documents retrieved in response to the user’s query \cite{lewis2021retrievalaugmentedgenerationknowledgeintensivenlp}. Thus, fine-tuning modifies the model itself, whereas RAG modifies the information provided to the model at the time of generation \cite{ovadia2024finetuningretrievalcomparingknowledge}.

    

\subsection{South Carolina Civil Rights History}


Over the past fifty years historians have developed a robust literature highlighting both the major themes and developments of the movement for Black Civil Rights as well as the local stories and issues that often do not figure in the national Civil Rights narrative. In 2005 historian Jacqueline Dowd Hall questioned the traditional periodization of the Civil Rights Movement and argued for a “long” Civil Rights Movement. According to the traditional timeline, Jim Crow laws implemented after Plessy v. Ferguson in 1896 led to a long period of quiet among Black leaders and activists. The movement then began in 1954 with the Brown v. Board school desegregation decision and continued through the Voting Rights Act of 1965 \cite{hall2005long}. These events serve as convenient beginning and end points for the Movement because they focus on the most obvious legal barriers to Black participation in modern economic and political life -- legal racial segregation and restrictions on the franchise. This traditional narrative also focuses heavily on the activities of Martin Luther King, Jr. and highlights events in limited southern locations. The role of women in the Movement, as well as the role of organizations such as the NAACP, are largely absent from this narrative. 

Given the limited geographic and temporal scope of the traditional narrative, the Civil Rights Movement in South Carolina remains opaque in the national consciousness. Many of the most significant legal challenges in the South occurred in South Carolina during the late 1940s, including several important cases that solidified Black participation in the Democratic Party primary after the Smith v. Allwright decision of 1944. These legal challenges include the Briggs v. Elliott school desegregation suit that formed the core of the Brown v. Board case \cite{lau2006democracy}. The post-1965 period saw the first legal challenge to the Voting Rights Acts in the form of South Carolina v. Katzenbach, one of the deadliest incidents of the entire Movement in the 1968 Orangeburg Massacre, and the final implementation of Brown v. Board fully sixteen years after the decision \cite[85--94]{minchin2011after,sellers2008orangeburg}. Although King visited South Carolina, he was far more active in other southern states. Local leaders were more integral to the fight in the Palmetto State, including women such as Modjeska Simkins and Septima Clark. Highlighting the events of the Movement in South Carolina therefore supports Hall’s contention and challenges the traditional narrative. 

South Caroliana Library at the University of South Carolina houses significant archival collections that attest to the Movement’s history in the state. Digital Collections in University Libraries has digitized many of these collections. They include the papers of Rev. Joseph DeLaine, who led the Briggs v. Elliott school desegregation lawsuit, activist and businesswoman Modjeska Simkins, newspaper publisher and Democratic Party leader John Henry McCray, NAACP leader and politician I. DeQuincey Newman, and newspapers such as McCray’s Lighthouse and Informer. These collections are indispensable to understanding the Civil Rights Movement in South Carolina. Without the information they contain,historians and the public would remain ignorant of aspects of the NAACP-led lawsuits of the 1940s, the White backlash of the 1950s, the protests and marches on the State House of the 1960s, and the violence of the late 1960s and early 1970s. These collections have, not surprisingly, served as the basis for dozens if not hundreds of scholarly monographs and peer-reviewed articles.

\subsubsection{Digital Collections Corpus Background}

Since the 1990s, academic libraries have been engaged in the digitization and online dissemination of their rare and unique manuscript collections, a movement catalyzed by the Library of Congress’s American Memory project \cite{saa_american_memory}. USC Libraries, which has made over a million items accessible online, has been digitizing its distinctive collections since 2004, facilitating research for scholars worldwide \cite{usc_digital_collections}. 
The Civil Rights Digital Collections are a small dataset of approximately 51,000 pages that includes transcripts and well-formed metadata concerning the Civil Rights Era. This dataset was selected because these collections have benefited from substantial effort, care, and resources, resulting in detailed metadata and reliable transcriptions. Furthermore, this dataset presents a unified theme, is vitally important to South Carolina history, and is unique. No other libraries hold these materials, and they are available digitally only through the University of South Carolina Libraries.

The dataset contains historical collections from the University of South Carolina’s South Caroliniana Library documenting civil rights organizing, Black journalism, school desegregation, religious leadership, and civic leadership in South Carolina from multiple, overlapping perspectives. In short, there are nine collections included in this dataset: South Carolina Council on Human Relations (SCCHR), John Henry McCray Papers, Modjeska Simkins Papers, Briggs v. Elliott Digital Collection, M. Hayes Mizell Papers, James T. McCain Papers, Eugene Avery Adams Papers, Joseph DeLaine Papers, and Isaac Samuel Leevy Papers. This paper focuses on the first dataset added, the John Henry McCray Papers.

\begin{figure}[!t]
    \centering
    \footnotesize

    \begin{tabular}{@{}c@{\hspace{0.04\linewidth}}c@{}}

        \begin{minipage}[t]{0.46\linewidth}
            \centering
            \includegraphics[width=\linewidth]{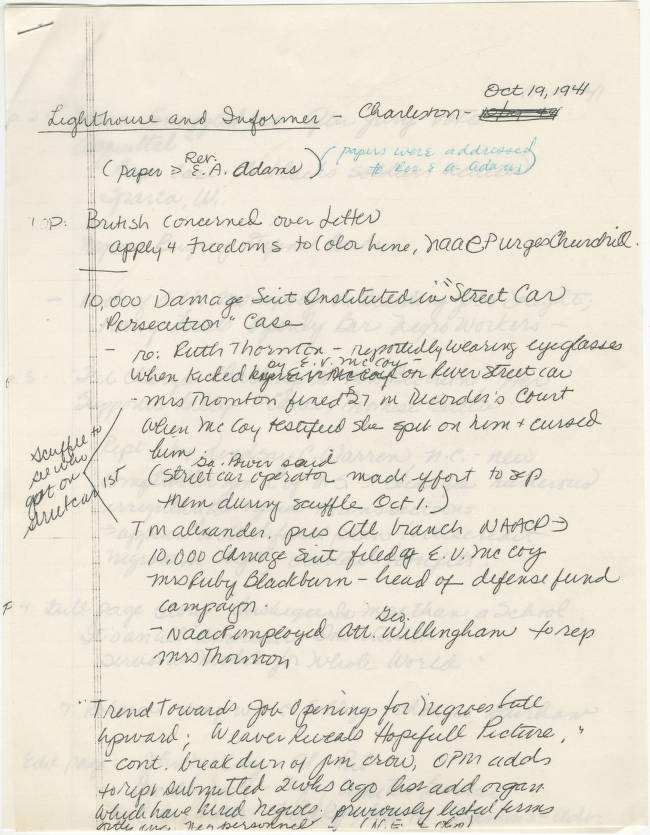}

            \vspace{0.35em}
            \textbf{(a) Handwritten document}
        \end{minipage}
        &
        \begin{minipage}[t]{0.46\linewidth}
            \centering
            \includegraphics[width=\linewidth]{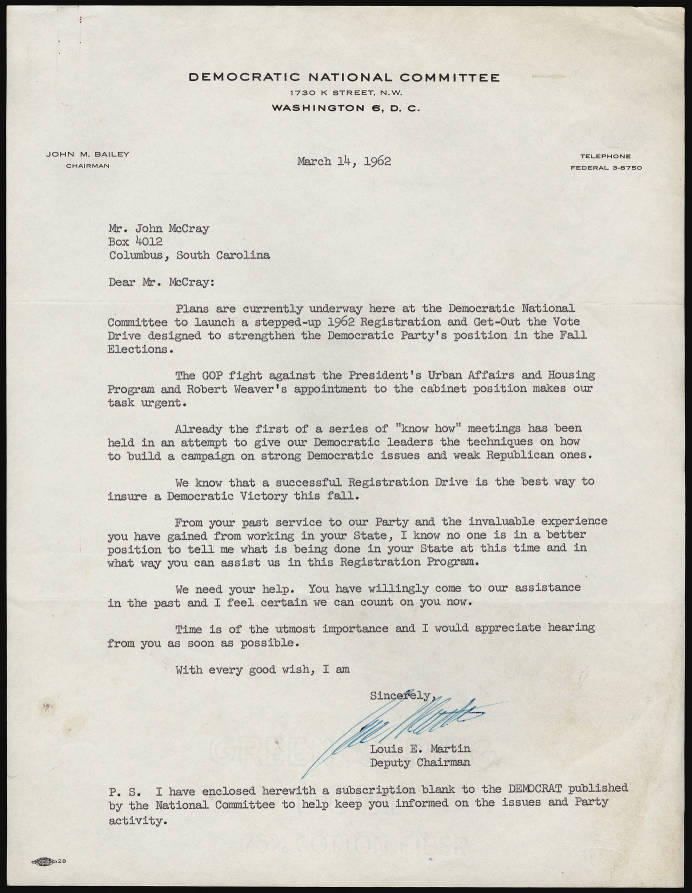}

            \vspace{0.35em}
            \textbf{(b) Printed document}
        \end{minipage}
        \\[0.8em]

        \begin{minipage}[t]{0.46\linewidth}
            \centering
            \begin{minipage}[t][9em][t]{\linewidth}
\begin{Verbatim}[fontsize=\tiny,
breaklines=true,
breakanywhere=true,
xleftmargin=0pt]
\\\\\\\\J^AAyuAJ- luaa-Ji - C$UZ^lt?terV) - Bryb
/<lt>?, jg^ I l-O: f \\\\\\\\s\\\\\\\\tf\\\\\\\\jhL fcul^d kjp^v1^^
taw 3&ia& cOaj jjj<gt>/ Cvum "foe OsLtf teaeqtuJ
5&*~ zpa &x ^th * Cuyi^- hP W k^oui^ud-
! wjA4rtis j*us*M<gt> w2^- /^-CU^i^ I <lt>
U V /
\end{Verbatim}
            \end{minipage}

            \vspace{0.2em}
            \textbf{(c) Transcript (handwritten)}
        \end{minipage}
        &
        \begin{minipage}[t]{0.46\linewidth}
            \centering
            \begin{minipage}[t][9em][t]{\linewidth}
\begin{Verbatim}[fontsize=\scriptsize]
Columbus, activity
\end{Verbatim}
            \end{minipage}

            \vspace{0.2em}
            \textbf{(d) Transcript (printed)}
        \end{minipage}

    \end{tabular}

    \caption{Document photographs and corresponding transcripts obtained with ABBYY. Transcripts highlight errors with (a) and (c) showing poor character recognition with handwritten documents, and (b) and (d) showing incompleteness with printed documents.}
    \label{fig:documents-transcripts}
\end{figure}

\subsubsection{Ground Truth and Trust}

The concept of ``ground truth'' is a foreign one to professional historians. Almost forty years ago, the historian Peter Novick despaired of historians agreeing on even the nature of “objectivity,” much less whether they could attain a ground truth on any historical topic \cite{novick1988noble}. Historians furthermore tend to see the advancement of historiography as {\bf dialectical}. That is, there may be multiple perspectives, even seemingly opposite or contradictory, which can all be true at the same time and hence, an {\em acceptable} answer may need resolving and synthesizing the tensions in these perspectives into a new, balanced response.
As George Santayana starkly stated in 1905, “[h]istory is always written wrong, and so always needs to be rewritten” \cite{santayana1921reason}.

Skepticism towards ground truths written by experts, and the need to study the same topics from different perspectives, therefore lie at the heart of historians’ professional training and activities. Historians are therefore predisposed to trust an impersonal chat less than ground truth statements of peer experts, even if those experts are anonymous. At the same time, professional historians will trust a chatbot to perform certain functions. Chatbots can perhaps provide linguistic insight regarding texts, insight that humans may miss, but professional historians will want to examine the original documents because professional standards require them to do so. 

However, when it comes to the general public, other factors may play a role. For example, users may prefer the convenience of lesser cognitive load from a guided single-result of a chatbot rather than reviewing and deciding from a list of results. Furthermore, they  may view results from a system more than a person, like a historian, due to {\em automation bias}\cite{automation-bias}. We believe that with  full-scale deployment of systems like \lh and detailed experiments, one can discover the true value  of chatbots for exploring history - our long-term goal. 

\subsection{ Trustworthy AI}

Advances in AI have increasingly pushed AI systems into the foreground of human-computer interaction, shifting their role from background decision-support tools towards core user-facing software products. Among these, LLM-based chatbots are the most prevalent. Their ability to respond to general informational needs and their natural language interface make them especially well suited for general consumer use. However, this widespread adoption raises questions of \emph{trust} in chatbots.

When a user interacts with an LLM-based chatbot like ChatGPT, Gemini, or Claude, the output is the result of a machine learning algorithm trained on vast swaths of text. The models themselves possess no cognitive agency or intent, leading some scholars to characterize these systems as ``stochastic parrots'' \cite{stochastic_parrots}. Users interacting with chatbots often rely on insufficient heuristics to guide their trust in these systems. Prior work indicates that the presence of citations in outputs significantly increases trust even when those citations are generated randomly \cite{ding2025citations}. Some cognitive scientists argue that the natural conversational interface that LLM-based chatbots enable also plays a role in earning misallocated trust when users anthropomorphize these systems \cite{GARRY20241078}.

Users' tendency to misallocate trust toward LLM-based chatbots is further complicated by limitations of the systems themselves. One issue is the lack of direct control and guarantees that LLMs provide to stakeholders. Prior work has noted that due to the nature of LLM post-training's large-scale crowd-sourced data, the resulting systems are not directly controlled by developers, instead being controlled by an amalgamation of influences from contributors to data, process, and development \cite{ouyang2022traininglanguagemodelsfollow}. A recent symptom of this lack of control is sycophancy, which is frequently attributed to biases in human feedback, where LLM-based chatbots will overly agree with users, even when wrong \cite{sharma2025understandingsycophancylanguagemodels}. Furthermore, LLMs have been shown to be non-deterministic even under settings which should provide determinism, a property which limits any guarantees that could be made about these systems \cite{llm_nondeterminism}.  These systems also have a tendency to \emph{hallucinate} information, providing plausible, but ungrounded or incorrect outputs \cite{bang2025hallulens, Huang_2025}. One solution to this, retrieval-augmented generation (RAG), which aims to ground LLM responses with external data, has been shown to decrease notions of safety in LLMs \cite{an-etal-2025-rag}.

Efforts in building trustworthy conversational assistants largely rely on one of two methods: normative, rule-based systems and model-centric alignment of generative models. Rule-based chatbots such as SafeChat and Rasa provide predictable behavior, enforce explicit do-not-answer policies, and maintain traceability to authoritative sources, making them well-suited for safety-critical domains such as healthcare \cite{SafeChat2025, bocklisch2017rasaopensourcelanguage, conversational_agents_in_healthcare, Martin_Pande_Schwander_Ajuwon_Pimmer_2024}. In contrast, recent work on LLM-based assistants focuses on improving safety through alignment techniques such as fine-tuning, reinforcement learning from human feedback, and principle-based constraints, often complemented by filtering and moderation \cite{ouyang2022traininglanguagemodelsfollow, bai2022constitutionalaiharmlessnessai}.

These two approaches towards creating and aligning trustworthy conversational assistants each provide unique advantages and disadvantages that are fundamental to their design. Recent work in AI has explored neuro-symbolic approaches to address and attend to both the limitations and strengths of these methods. We explore the use of SOFAI, a cognitive architecture based on dual-process theory, through SafeGenChat, and implementation of SOFAI in the domain of trustworthy conversational assistants \cite{BergamaschiGanapini2025, safegenchat}.

We also note that although mainstream AI and IR systems provide 
single results, and \lh is along the same trend, an active area in  AI is looking at how to support multiple, nuanced perspectives  \cite{nguyen2026positionvince-ai-practicalalignment}. We see this as a promising avenue of future work to support historians' dialectical expectations. 

\section{Problem}
\label{sec:problem}



In building a trustworthy conversational assistant for exploring digitized historical archives, we consider an information-seeking scenario in which a user interacts with a system to obtain information on a topic of interest. While traditional information retrieval (IR) systems, such as keyword-based search, assume well-specified queries, conversational interfaces allow information access to occur within a broader dialog context.

Prior work on collaborative IR characterizes such settings as interactive processes in which information is exchanged over multiple turns, potentially incorporating conversational context and system memory \cite{conversation-ir-framework}.  In this work, we focus on enabling flexible, context-aware, generative access to information while ensuring that responses remain anchored in validated knowledge through symbolic or rule-based mechanisms.

Let $U$ denote the set of possible user utterances. Given a user utterance $u \in U$, the system has access to two knowledge sources. The first is a validated question-answer dataset,
 $$D_E = \left\{ (q_i , a_i , s_i )\right\}_{i=1}^n,$$
where each $q_i$ is a question about the archive, $a_i$ is an answer created or reviewed by domain experts, and $s_i$ denotes supporting source information or provenance. This dataset represents high-trust archival knowledge, but its coverage is limited to questions anticipated during expert annotation..

The second knowledge source is the archival corpus,
$$D_A = \left\{d_j\right\}_{j=1}^m,$$
where each $d_j$ is an archival document, transcript, metadata record, or other textual representation of a digitized primary source. This corpus represents a broader, but less structured base of knowledge. It may contain evidence related to user queries which are not anticipated during expert annotation.

Given a user utterance $u$, the system must select an appropriate strategy. 
If $u$ can be mapped to a question or scenario represented in $D_E$, the system should prefer the corresponding expert-validated answer.
If $u$ is not answerable by $D_E$, the system may attempt to answer using relevant evidence retrieved from $D_A$.

This formulation creates a tradeoff between trust and coverage. A system that relies only on $D_E$ can provide controlled and reliable answers, but only for a limited number of user utterances. A system that relies only on $D_A$ can respond to a wider range of questions, but lacks the stronger reliability guarantees provided by expert-vetted answers. The goal therefore is to design a system that can balance these sources.


\section{Approach}
\label{sec:approach}

The following sections describe, in depth, our approach to creating these chatbots. In summary, the process consists of two main steps: first, data curation, described in Sections \ref{sec:data_processing} and \ref{sec:approach-ground-truth}, which includes both transcribed primary documents and expert-created answers; and second, solver development, described in Section \ref{sec:approach-ai}, which focuses on building systems that can respond to user queries.

\subsection{Dataset Selection}

While the broader Civil Rights-era dataset contains nine digitized collections from the University of South Carolina Libraries, the chatbot in this project is trained specifically on the John Henry McCray collection. This collection was chosen from the larger set because of its size, historical significance, and strong connections to the other Civil Rights-era collections described in Section \ref{sec:related-work}. The McCray collection originally consists of 19,329 machine-generated transcripts. Before transcript cleaning, these transcripts contain approximately 6.75 million tokens; after cleaning, they contain approximately 5.68 million tokens. Including metadata, the cleaned McCray corpus contains approximately 6.46 million tokens. Table \ref{tab:token-counts} compares the size of the McCray collection with other entries in the Civil Rights Digital Collections.

The McCray collection provides a focused and sufficiently large foundation for chatbot training. John Henry McCray was associated with South Carolina’s Civil Rights movement, particularly through his work in Black journalism and activism. His papers also overlap in subject matter, people, and organizations with other collections in the broader dataset, allowing the chatbot corpus to remain focused on one collection while still reflecting parts of the larger Civil Rights-era archival context.



\subsection{Data Processing and  Cleaning }
\label{sec:data_processing}


\begin{table}[t]
\centering
\scriptsize
\setlength{\tabcolsep}{3pt}
\renewcommand{\arraystretch}{1.2}
\caption{Token counts (Llama-3.1) by dataset before and after transcript cleaning. Combined indicates Transcripts plus Metadata}
\label{tab:token-counts}

\resizebox{\columnwidth}{!}{%
\begin{tabular}{lrrrrr}
\hline
\textbf{Dataset}
& \makecell{\textbf{Original} \\ \textbf{Transcript}}
& \makecell{\textbf{Cleaned} \\ \textbf{Transcript}}
& \textbf{Metadata}
& \makecell{\textbf{Combined} \\ \textbf{(Original)}}
& \makecell{\textbf{Combined} \\ \textbf{(Cleaned)}} \\
\hline
eaadams & 368,464 & 361,912 & 97,306 & 465,770 & 459,218 \\
mccray  & 6,746,949 & 5,675,198 & 788,388 & 7,535,337 & 6,463,586 \\
scchr   & 5,642,154 & 5,343,116 & 496,474 & 6,138,628 & 5,839,590 \\
simkins & 7,359,033 & 6,388,465 & 215,759 & 7,574,792 & 6,604,224 \\
\hline
\end{tabular}%
}
\end{table}



The John Henry McCray papers were initially transcribed with ABBYY FineReader in 2018.  These transcripts covered 19,504 documents, some of which contained descriptions from domain experts. Since many of the McCray papers are handwritten documents, these transcripts had significant errors that had to be addressed before using them in training. Many optical character recognition (OCR) systems are not designed for use on handwritten documents, especially cursive, leading to many transcripts having substantial errors. An example of which is shown in Figure \ref{fig:documents-transcripts}a. Furthermore, even printed documents can produce errors. Figure \ref{fig:documents-transcripts}b shows a printed document where the generated transcript contains only two words. Such documents where most of the content is missing provide no useful information in training and are thus excluded.

Recent advances in OCR systems and document AI have increasingly shifted toward Transformer- and vision-language-based systems. Examples of this work include TrOCR and DeepSeek-OCR \cite{trocr, deepseekocr}. However, we retain the existing OCR transcripts for this project rather than generating new transcripts with vision-language models (VLMs) or multi-modal large language models (MLLMs). In our pilot tests on handwritten historical documents, MLLM-generated transcripts were often fluent and broadly accurate, but were less reliable when identifying proper nouns such as personal names, organization names, and place names. These observations are consistent with prior work in MLLM-based OCR on handwritten documents \cite{gutteridge2026judgebookcoverinvestigating}. This is a significant concern for historical datasets because such errors can introduce false factual claims; in this context, a loss of information is preferable to the fabrication of information. We therefore do not use VLM or MLLM-generated transcripts as a primary source; instead, we focus on cleaning, correcting, and validating the pre-existing OCR transcripts.

We approach cleaning in two phases: an initial filtering stage, where transcripts deemed unusable are removed, and an error-correction stage, where OCR artifacts are repaired. In the first stage, we filter out records with no transcript, duplicate transcript text, and transcripts with too little recognizable language to be useful. In the second stage, we correct common OCR errors, remove non-text artifacts, normalize spacing and formatting, and apply targeted substitutions for recurring misread words or characters. This produces a cleaner set of transcripts while preserving documents that are likely to contain usable textual information.

This process is supported by a custom NLTK vocabulary derived from document titles and descriptions, as well as a domain-specific Spello correction corpus built from Wikipedia articles related to Southern and civil rights history \cite{nltk, spello, wikipedia}.

\subsection{Creating a Gold Standard for Evaluation}
\label{sec:approach-ground-truth}

\begin{table*}[t]
\centering
\scriptsize
\setlength{\tabcolsep}{3pt}
\renewcommand{\arraystretch}{1.3}
\caption{Comparison of short and long responses to the question: ``Was it necessary for John McCray to organize the Progressive Democratic Party?''}
\label{tab:respondent-comparison}

\begin{tabularx}{\textwidth}{
@{}
p{1.8cm}
>{\raggedright\arraybackslash}X
>{\raggedright\arraybackslash}X
>{\raggedright\arraybackslash}X|
>{\raggedright\arraybackslash}X|
>{\raggedright\arraybackslash}X
@{}
}
\hline
& \textbf{Resp. 1}
& \textbf{Resp. 2}
& \textbf{Resp. 3}
& \textbf{Aggregated}
& \textbf{Evaluation} \\
\hline

\textbf{Short Answer}
& Yes.
& The South Carolina Democratic Party operated as a private club, even after the \emph{Smith v. Allwright} decision, and did not allow Black participation either in its conventions or in its primaries.
& Not necessary, but effective.
& Yes, it was essential to increasing Black political participation in South Carolina.
& Accuracy - 3; Comprehensiveness - 1; Readability - 5. 

Although the answer is correct as far as it goes, it is far too short to be useful to those seeking information on this topic. \\

& & & & & \\[-0.6em]

\textbf{Long Answer}
& Yes, John McCray helped create the PDP because Black people were not allowed to join the State Democratic Party. 
& The South Carolina Democratic Party operated as a private club, even after the Smith v. Allwright decision, and did not allow Black participation either in its conventions or in its primaries. The Progressive Democratic Party was necessary as a pressure group so that the National Democratic Party would recognize the unfairness of South Carolina Democrats' treatment of Black voters.  
& Creating the party and challenging the White Democratic grip on South Carolina accomplished at least two main goals: 1) the party galvanized Black voters' participation in politics in the 1940s and 1950s 2) the party's work at the national convention pressured the national party to acknowledge African American voters and pressured the state party for change 3) developed a court case, Elmore v. Rice, that successfully challenged the all-white primary system. 
& It was necessary for John McCray to help organize the Progressive Democratic Party because the South Carolina Democratic Party (PDP) operated as a private club that prohibited Black participation in its conventions and primaries. Thus, the PDP was created to provide Black South Carolinians with opportunities for political participation, and to expose the official state party's discriminatory practices. Additionally, the work of the PDP also helped to organize the court case Elmore v. Rice, which successfully challenged South Carolina's all-white primary system. 
& Accuracy - 5; Comprehensiveness - 4; Readability - 5.

This answer is comprehensive and accurate. I believe that as more documents are added, this answer will become even more robust.  \\

\hline
\end{tabularx}
\end{table*}

\begin{figure}
 \centering
  \includegraphics[width=0.4\textwidth]{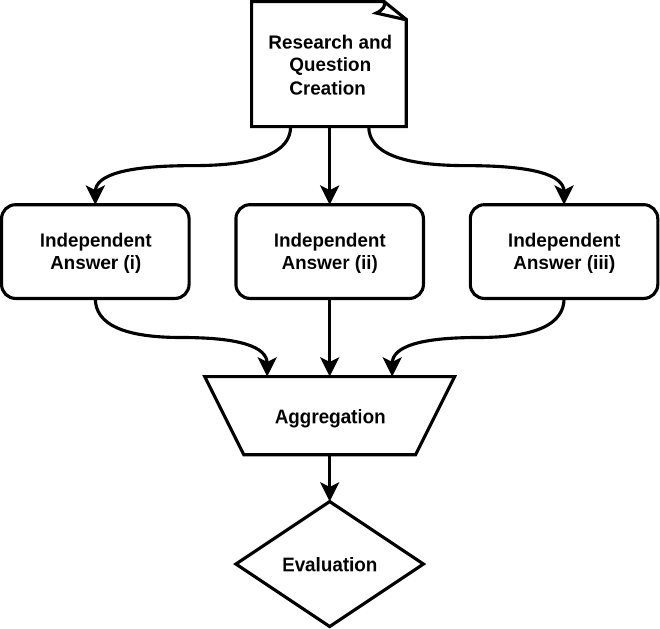}
  \caption{Process for determining {\em "gold-standard"} responses for questions with subjective answers. We rely on experts in the roles of independent respondents (3), aggregator and evaluator - total 5 - to synthesize an acceptable answer.}
  \label{fig:chatbot-llm-subjective-eval-process}
\end{figure}

A core component to the development of \lh has been creating a \emph{gold standard} for responses to user queries. We define this gold standard as answers to questions related to the collections, which have been created and evaluated by subject matter experts. We create these through a four-stage process: question curation, expert annotation, aggregation, and evaluation. This four-stage process, shown in Figure \ref{fig:chatbot-llm-subjective-eval-process}, aims to ensure that the resulting answers are accurate, unbiased, and adequate for the queries. 

For each collection, we create a set of \emph{scenarios}, representing questions one might ask about important figures, entities, or events that are covered in the collection. We define each scenario as ten questions that cover this topic, with a further distinction of five objective and five subjective questions, with the former being questions that have one clear factual answer and the latter being questions that require nuanced responses and may have several answers.

We begin the creation of these scenarios by finding important subjects within each collection. Example subjects from the John Henry McCray collection are: political initiatives of McCray, the NAACP, a political organization created to fight for the equal rights of Black people beginning in the early 20th century, the Lighthouse and Informer, a Black newspaper run by John McCray from 1938 to 1954,  and the Progressive Democratic Party, a party co-founded by John McCray in 1944 which aimed to increase Black political participation in South Carolina.

After the scenarios are chosen, individual questions are created about each topic. The questions are designed so that they can either be fully or at least partially answered by the dataset. Because we have little control over which sources domain experts draw from to answer these questions, this measure does not guarantee that the end product, a gold-standard answer, will align with the information in the documents.

Once the scenarios are created, the questions are shared with three graduate students of history who are familiar with South Carolina Civil Rights movement. They answered these questions based on their knowledge and the McCray papers. The three independent responses were aggregated by a history scholar. The aggregated responses were finally evaluated by an expert based on three criteria: \emph{accuracy}, \emph{comprehensiveness}, and \emph{readability} (Figure 3). The aggregation and evaluation processes are performed by different scholars to avoid potential bias toward their own versions of the aggregated answers.

\subsection{AI Approach for Dialog Management }
\label{sec:approach-ai}




\lh is implemented using the 
SafeGenChat architecture \cite{ safegenchat} -- an approach where a range of {\em dialog managers} (DMs - also called solvers) are available, the chatbot subcomponent that deals with deciding how to respond to a user utterance. \lh can be configured to use only the learning-based DM (also called S1), the rule-based DM (also called S2), or a hybrid DM which can invoke both. The hybrid DM chooses which solver to call dynamically at each dialog stage learning a policy based on the history of  past interactions and considerations like risks.

\subsubsection{LLM-based}
\label{sec:approach_llm}
We refer to the specific learning based model used here as M1-1. The model was initialized from Meta’s Llama-3.1-8B checkpoint, a lightweight pre-trained model \cite{grattafiori2024llama3herdmodels}. We fine-tuned the model on the McCray collection using LoRA \cite{hu2021loralowrankadaptationlarge} for 10 epochs with a learning rate of $2 \times 10^{-5}$, batch size of 2, LoRA rank $r=16$, LoRA $\alpha=32$, and dropout of 0.1. Training used the AdamW optimizer with weight decay 0.01, a cosine learning-rate schedule, and 100 warmup steps. We then performed a lightweight supervised fine-tuning stage on a randomly selected subset of Google Natural Questions \cite{kwiatkowski-etal-2019-natural} totaling 2,000 question-answer pairs using LoRA rank $r=8$, LoRA $\alpha=16$, dropout of 0.1, a learning rate of $5 \times 10^{-5}$, batch size of 2, and gradient accumulation over 4 steps for 2 epochs. This second stage used the same AdamW optimizer, cosine learning-rate schedule, weight decay of 0.01, and 100 warmup steps.

We use two other LLM configurations to baseline performance: M1-2 and M1-3. We employ Llama-3.1-8B-Instruct (M1-2), an instruct-tuned variant of the base model we use in our main LLM-based method. In order to isolate differences in semantic content and eliminate noise introduced by stylistic variations in model outputs, we use a fine-tuned version of Llama-3.1-8B using the same Google NQ training described above. This fine-tuned variant is referred to as M1-3. These two baselines are chosen to provide: (a) a more generally capable, but stylistically different model and (b) a similarly narrow, but stylistically similar model. All models are evaluated under greedy decoding. This choice is deliberate and influenced by the sensitivity of the domain in which these chatbots are deployed.

All models are evaluated under greedy decoding to reflect the most likely answer they give when queried.

\subsubsection{Rule-based}

We develop a rule-based variant of \lh, which will be referred to as M2, using SafeChat \cite{SafeChat2025}, a Rasa-based framework for building source-grounded chatbots. We configure the system using the default SafeChat settings and train it on the McCray scenario questions introduced in our gold-standard evaluation dataset. We refer to this rule-based model as M2.

\subsubsection{Hybrid}



We develop a hybrid variant of \lh, referred to as M3, using SafeGenChat \cite{safegenchat}, which combines generative and rule-based components. In this configuration, Llama-3.1-8B-Instruct (M1-2) serves as the System 1 model, while SafeChat serves as the System 2 solver. We instantiate the system with default thresholds and parameters, along with a keyword-based risk model that assigns varying levels of conversational risk to terms related to civil rights history.

The hybrid workflow can be summarized as follows. First, the system receives a user query and passes it to System 1, which generates an initial response. The system then evaluates the perceived risk of the query together with its confidence in the System 1 response. When the query is judged to be low risk and the response sufficiently reliable, the System 1 answer is returned. When the query is higher risk or the System 1 response is uncertain, the system instead falls back to System 2 to provide a more controlled response.

\begin{table}[t]
\centering
\caption{Model shorthands used in Lighthouse.}
\label{tab:model-shorthands}

\renewcommand{\arraystretch}{1.25}
\begin{tabularx}{\linewidth}{@{} l X X @{}}
\toprule
\textbf{Shorthand} & \textbf{Type} & \textbf{Description} \\
\midrule
M1-1 & Fine-tuned LLM & Llama-3.1-8B fine-tuned on McCray and Google NQ data. \\
M1-2 & Instruct LLM baseline & Llama-3.1-8B-Instruct baseline. \\
M1-3 & Fine-tuned LLM baseline & Llama-3.1-8B fine-tuned on Google NQ data. \\
M2 & Rule-based & SafeChat-based rule-based Lighthouse. \\
M3 / H1 & Hybrid & SafeGenChat combining Llama-3.1-8B-Instruct and SafeChat. \\
\bottomrule
\end{tabularx}
\renewcommand{\arraystretch}{1}

\end{table}
\section{Evaluation}
\label{sec:eval}



Our automated evaluation is organized around the nature of questions: objective or subjective. We will refer to them as AE1 and AE2 respectively. These evaluations are performed under six textual comparison metrics: Jaccard similarity, ROUGE-L, Jensen-Shannon Divergence, BLEU, Levenshtein distance, and BERTScore, which are all normalized to be similarity scores. These evaluations serve to answer a primary research question: How well can these models answer both objective and subjective questions about digital collections?



For AE1, we evaluate all models on 50 objective questions. The objective questions are drawn from all scenarios created for the McCray collection: \textit{McCray}, \textit{NAACP}, \textit{Lighthouse and Informer}, \textit{PDP}, \textit{Black Journalism}, \textit{Voting Rights}, \textit{Education}, \textit{Civil Rights Leaders}, \textit{Jim Crow}, and \textit{Racial Progress}. These questions do not undergo the process covered in Section \ref{sec:approach-ground-truth}. 

For AE2, we evaluate all models on 10 subjective questions that are drawn from the scenarios \textit{McCray} and \textit{NAACP}. Accuracy is reported as textual comparison to the short form of the gold-standard reference responses. Because of the small sample size, we opt to do comparisons against the short version of answers to reduce noise that may be introduced in comparisons against longer responses.

Results for AE1, shown in Figure 4, reflect the nature of these chatbots. M2, which directly retrieves known answers, performs best, followed by M3, which combines direct retrieval with LLM paraphrasing. Finally, M1-1 outperforms both M1-2 and M1-3, the baseline LLM-based approaches, indicating that the training on the McCray collection is indeed improving response accuracy. Low scores for M1-2 are likely due to answer length as the reference responses are often only a few words, while the observed outputs from M1-2 often spanned multiple sentences.

In AE2, we similarly see high performance from both M2 and M3, but see M1-2 outperforming both M1-1 and the M1-3 baseline. Both M1-1 and the M1-3 baseline are trained to give concise and direct answers as seen in the Google NQ corpus, while M1-2 gives full-length answers to queries that match the style of our subjective-question dataset. These findings highlight a key issue with LLM evaluation on tasks that require long-form answers: stylistic differences in output can often influence results more than content.





\begin{figure}
    \centering
    \includegraphics[width=1.0\linewidth]{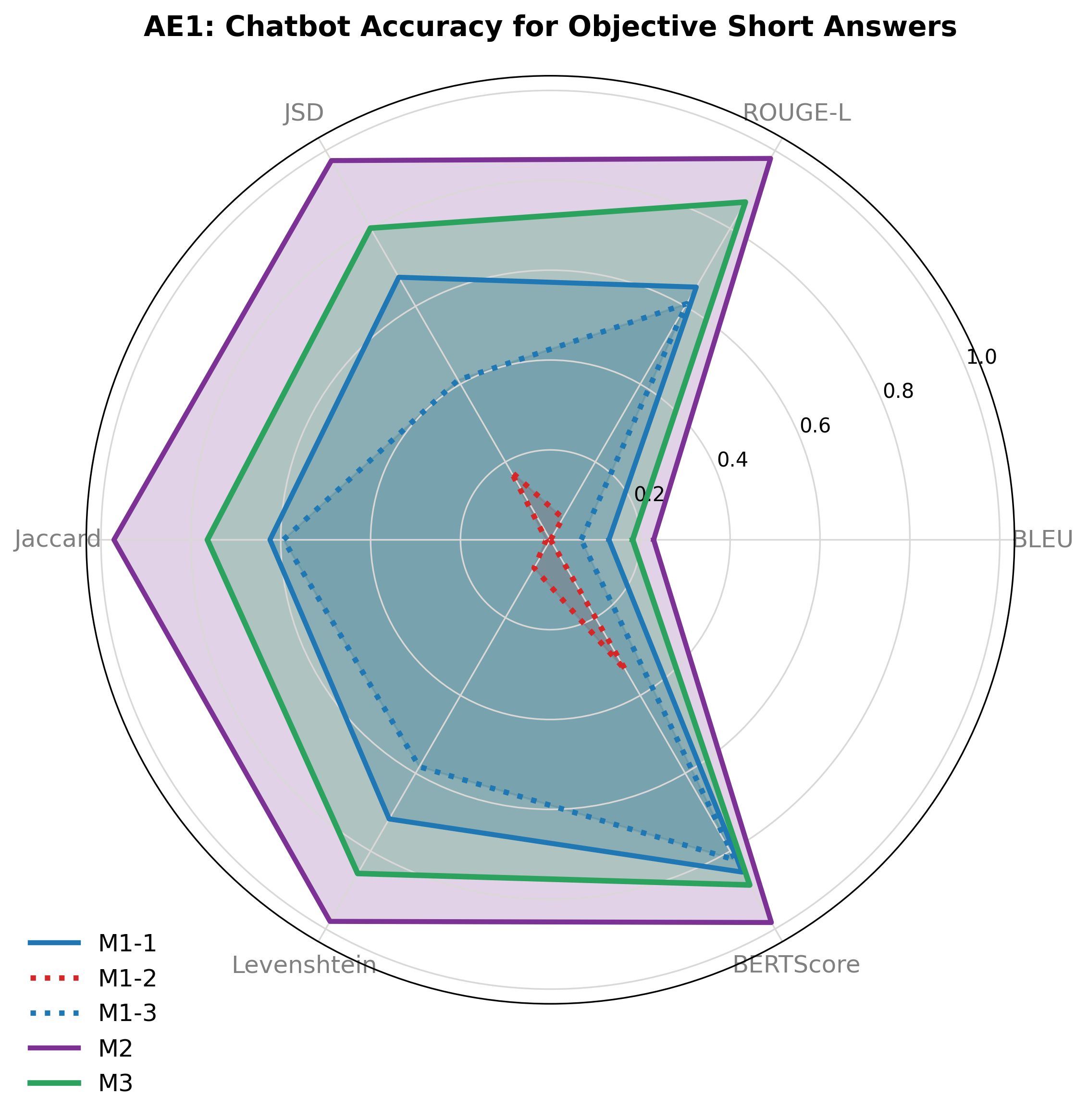}
    \caption{Chatbot accuracy for AE1. Larger areas indicate better performance. Dotted lines indicate baseline approaches.}
        \label{fig:ae1}
\end{figure}


\section{Discussion }
\label{sec:discussion}


We now discuss the presented work for its significance, limitations and avenues for future work.

\subsection{Societal Significance}

Given the nature of historiographical orientation and training for college undergraduates, instructors are likely to incorporate chatbots such as this into a broader orientation on the secondary literature on a given topic as well as lessons on primary source and media literacy. Instructors are most likely to provide students with the primary sources that form the basis of the large language model and ask students to evaluate the chatbot’s responses against the sources. They will almost certainly also ask students to compare responses to scholarly interpretations presented in the secondary literature on the relevant topics. Instructors in history departments would rightly, in light of their professional training and standards, reject the use of chatbots as the sole or even primary interpreter of the primary sources that form their basis nor would they favor their interpretations over those of professional historians.

The chatbot can nevertheless introduce students to primary sources as digitized archival materials are often housed within systems whose search functions are not intuitive and that may require extra training to navigate. Furthermore, as long as the primary resources are available through the chatbot and connected to its responses, this platform could also assist general audiences in finding materials in digital archives.




\subsection{Technical Significance}


The technical significance of \lh lies in its neuro-symbolic design for conversational access to sensitive historical archives. Purely generative systems offer flexibility, but provide limited guarantees of accuracy, provenance, and consistency. Purely rule-based systems offer stronger control and traceability, but only for questions anticipated in advance. \lh combines these approaches, using generative capabilities for flexibility while relying on symbolic mechanisms to prioritize validated answers, cite sources, and avoid unsupported responses.

\lh also provides an evaluation framework suited to historical archives. By separating objective questions from subjective ones, the evaluation reflects the different kinds of knowledge users seek from archival collections. Objective questions can be evaluated against factual answers, while subjective questions require expert judgment, aggregation, and review.


\begin{figure}
    \centering
    \includegraphics[width=1.0\linewidth]{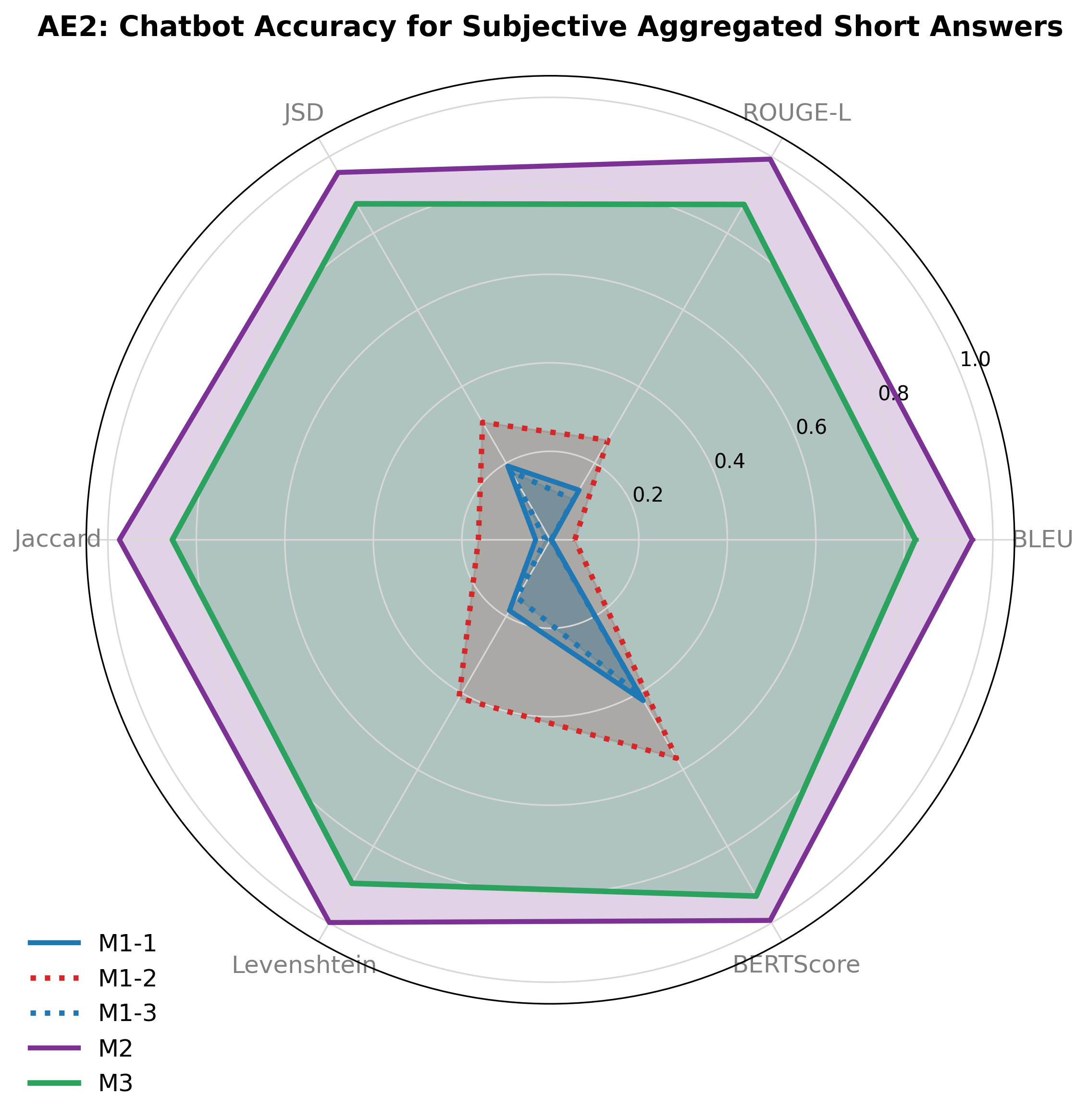}
    \caption{Chatbot accuracy for AE2. Larger areas indicate better performance. Dotted lines indicate baseline approaches.}
    \label{fig:ae2}
\end{figure}

\subsection{Limitations \& Future Work}

This work represents a narrow exploration of how AI can further improve access to historical archives. We examine three approaches: a learning-based approach using fine-tuned LLMs, rule-based with a traditional rule-based chatbot, and a hybrid approach, combining the two. These approaches, while covering the primary methods for creating chatbots, are not a complete representation of efforts in chatbot development or information retrieval. Some notable efforts such as creating domain-specific foundational models, specialized retrieval models for RAG, and even agentic workflows were not explored. These more advanced approaches may yield better results.

Another limitation is the need for expert-answered questions in both the rule-based and hybrid approaches. These methods work well to answer questions that have already been addressed and provide a reliable means of doing so, but do not enable exploratory information retrieval.

\section{Conclusion}
\label{sec:conclusion}

This paper presented \lh, a conversational assistant for exploring Civil Rights archival collections with attention to trustworthiness, provenance, and the reduction of user harm. Using the John Henry McCray collection as an initial case study, we compared generative, rule-based, and hybrid approaches and developed an expert-informed evaluation process for both objective and subjective historical questions. The results suggest that while LLMs can support more flexible access to archival materials, that may supplement search-based access, trustworthy use in historical domains requires {\bf hybrid mechanisms} for source grounding and expert validation. \lh therefore represents an early step toward conversational archival systems that can broaden access to complex collections while preserving the role of evidence, context, and human expertise.


\section*{Acknowledgment}
\label{sec:ack}

The authors gratefully acknowledge AspireAI grants by USC's AspireAI grants to 
Lighthouse No. 300000-25-72049) and  SafeGenChat (No. 155900-25-72050). They are also thankful for the computational resources provided by the Theia high performance computing cluster at USC supported by NSF Grant No. 2320292, and by USC's Research Computing  (RRID:SCR\_027488).

\bibliographystyle{IEEEtran}
\bibliography{references}

\end{document}